\documentclass[10pt,twocolumn,letterpaper]{article}

\usepackage[pagenumbers]{iccv} 
\usepackage{textcomp}
\usepackage{stfloats}
\usepackage{url}
\usepackage{verbatim}
\usepackage{graphicx}
\usepackage{multirow} 
\usepackage{adjustbox}
\definecolor{iccvblue}{rgb}{0.21,0.49,0.74}
\usepackage[pagebackref,breaklinks,colorlinks,allcolors=iccvblue]{hyperref}

\title{MCAM: Multimodal Causal Analysis Model for Ego-Vehicle-Level \\Driving Video Understanding}

\author{
	Tongtong Cheng\thanks{Equal contribution}$\;$$^,$$^1$,
	Rongzhen Li$^\dag$$^,$$^2$,
	Yixin Xiong$^*$$^,$$^1$,
	Tao Zhang$^1$,
	 Jing Wang$^3$, 
	 and Kai Liu\thanks{Corresponding author}$\;$$^,$$^1$$^,$$^2$ \\
	$^1$Department of Computer Science, Chongqing University, China \\
	$^2$National Elite Institute of Engineering, Chongqing University, China \\
 	$^3$College of Computer Science and Technology, National University of Deffense Technology, China\\
	{\tt\small \{totcheng0105,1xin,ztao0910\}@stu.cqu.edu.cn wangjing@nudt.edu.cn \{lirongzhen,liukai0807\}@cqu.edu.cn}
}

\begin{document}
\maketitle
\label{sec:intro}
\begin{abstract}
Accurate driving behavior recognition and reasoning are critical for autonomous driving video understanding. However, existing methods often tend to dig out the shallow causal, fail to address spurious correlations across modalities, and ignore the ego-vehicle level causality modeling. To overcome these limitations, we propose a novel Multimodal Causal Analysis Model (MCAM) that constructs latent causal structures between visual and language modalities. Firstly, we design a multi-level feature extractor to capture long-range dependencies. Secondly, we design a causal analysis module that dynamically models driving scenarios using a directed acyclic graph (DAG) of driving states. Thirdly, we utilize a vision-language transformer to align critical visual features with their corresponding linguistic expressions. Extensive experiments on the BDD-X, and CoVLA datasets demonstrate that MCAM achieves SOTA performance in visual-language causal relationship learning. Furthermore, the model exhibits superior capability in capturing causal characteristics within video sequences, showcasing its effectiveness for autonomous driving applications. The code is available at \textcolor{red}{\textit{\url{https://github.com/SixCorePeach/MCAM}}} \end{abstract}    
\section{Introduction}

With the advancement of autonomous driving technologies, understanding driving videos has become a critical research area~\cite{marcu2024lingoqa, liu2023cross, jin2023adapt, yuan2024rag}. This is due to the cost-effective acquisition of videos, which provide richer feature information than images~\cite{chen2025h,khan2024road}. Moreover, understanding driving videos is crucial for comprehending autonomous driving behavior, supporting the wider adoption of intelligent vehicles.
Recent studies have explored vision-to-language methods~\cite{du2024reversed, xu2024drivegpt4,chen2024egocentric, islam2024video, gu2023text}, which translate driving behaviors from visual data into high-level textual narratives. However, these methods~\cite{fan2024mllm,wang2024rac3,aafaq2019spatio} primarily focus on simple behaviors, neglecting the causal relationships between driving actions and their environmental context. Furthermore, the learning process is often hindered by spurious correlations, which could severely degrade the accuracy of causal analysis. Thus, developing a robust model with causality that mitigates spurious correlations is essential for advancing autonomous driving technologies.
\begin{figure*}[htbp]  
	\centering  
	\includegraphics[width=0.8\textwidth]{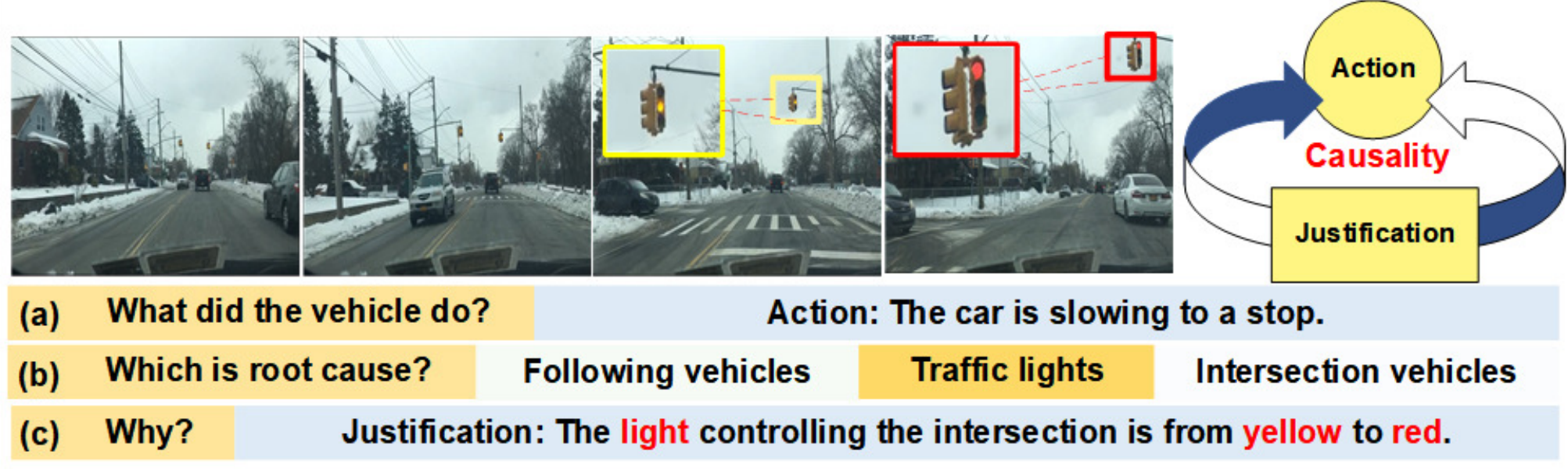}  
	\caption{The relationship of justification and action exists in ego-vehicle  level driving video understanding. In the initial state, the vehicle moves along the main road on the left side. There is a clear road with a visible traffic light in the distance. As the traffic light turns from yellow to red, the vehicle slows down and stops, preventing a violation of traffic regulations and reducing the risk of an accident. The yellow box and red box are added by manually to support the observation results.}  
	\label{fig1}  
\end{figure*}

Current approaches to driving behavior understanding can be categorized into two main streams: caption generation~\cite{tang2021clip4caption, zhang2024vision} and retrieval-based matching~\cite{madasu2024icsvr, ma2023llavilo, xu2024retrieval, xiao2024bridging, moon2023query}. For instance, Jin et al.~\cite{jin2023adapt} proposed ADAPT, a Video Swin Transformer-based architecture~\cite{lin2022swinbert}, which pairs driving behavior descriptions with comprehensive predictions. Similarly, Fan et al.~\cite{fan2024mllm} introduced MLLM-SUL, a framework combining multi-scale feature extraction with large language models (LLMs) like LLaMA for cross-modal video understanding. As illustrated in Figure \ref{fig1}, such methods establish causal links between actions (e.g., stopping) and environmental changes (e.g., traffic lights turning from yellow to red). While these approaches provide insights into driving behaviors, their causal reasoning is often limited to probabilistic correlations rather than detailed causal inference. On the other hand, retrieval-based methods, such as RAG-driver~\cite{yuan2024rag} and RAC3~\cite{wang2024rac3}, leverage retrieval mechanisms to enhance scene understanding by searching for similar scenarios. However, these methods heavily depend on large-scale datasets and fail to effectively mitigate the problem of spurious correlations inherent.

To reduce the detrimental impact of spurious causal relationships—stemming from spatial-temporal correlations in driving video understanding—and to establish authentic causal relationships, researchers have focused on a sophisticated realm of causal learning that emphasizes attention mechanisms~\cite{lo2023spatio} and causal reasoning~\cite{zhu2024causal, wang2023deconfounding, wang2022weakly}, with particular emphasis on exploring and understanding spatio-temporal causal relationships. For example, Cong et al.~\cite{cong2021spatial} introduced spatio-temporal attention to capture intra-frame and inter-frame relationships. Chen et al.~\cite{chen2024llcp} proposed the LLCP, which segments vehicle states into environmental context, neighboring vehicle's behaviors, and historical states. In addition, Liu et al.~\cite{liu2023cross} proposed the CMCIR framework, utilizing causal intervention techniques to facilitate event-level comprehension in driving scenarios. Despite their significant advancements, these approaches fail to adequately model the dynamic state transitions of causal entities, particularly overlooking ego-vehicle behavior understanding and the establishment of causal relationships.

To address this gap, this paper introduce the Multimodal Causal Analysis Model (MCAM), a framework designed for comprehensive driving video understanding. MCAM employs a Driving State Directed Acyclic Graph (DSDAG) to decompose complex driving behavior into discrete states, enabling fine-grained causal reasoning and dynamic interaction modeling. The framework consists of three components: (1) a Multi-level Feature Extractor (MFE) captures local and global features from videos; (2) a Causal Analysis Module (CAM) constructs causal relationships using DSDAG; and (3) a Vision-Language Transformer (VLT) integrates these features and relationships to generate detailed descriptions and reasoning.
The main contributions are as follows:
\begin{itemize}
	\item \textbf{Driving State Directed Acyclic Graph (DSDAG).} We propose Driving State Directed Acyclic Graph (DSDAG) to model dynamic driving interactions and state transitions, providing a structured theoretical foundation for the Causal Analysis Module (CAM).
	\item \textbf{Multimodal Causal Analysis Model (MCAM).} We propose a causality-analysis ego-vehicle level driving video understanding framework, named Multimodal Causal Analysis Model (MCAM), to discover true causal relationships through driving state modeling and achieve robust ego-vehicle level driving video understanding performance. To the best of our knowledge, we are the first to introduce causal analysis structures for ego-vehicle level driving video understanding tasks.
	\item \textbf{Comprehensive Evaluation.} Extensive experiments on the BDD-X and CoVLA datasets show the effectiveness of our MCAM for Mitigating the interference of spurious correlations to achieve robust and promising ego-vehicle level driving behavior understanding.
\end{itemize}

\section{Related Work}
\subsection{Causal Representation}
Building on the structural causal model proposed by Pearl et al.\cite{pearl2009causality}, researchers have further explored effective causal modules. 
For instance, Yang et al.\cite{yang2021causal} introduced a causal attention mechanism that leverages front-door adjustment to enhance the causal learning capabilities of transformer architectures. 
Additionally, Wei et al.\cite{wei2023visual} developed a causal scene separation module to distinguish between causal and non-causal visual scenes. They also proposed a question-guided refiner to select consecutive video frames for modeling causal relationships. 
Furthermore, Rahman et al.\cite{rahman2024modular} presented the Modular-DCM algorithm, which integrates causal structures with pre-trained conditional generative models to effectively eliminate the influence of confounding variables and accurately estimate causal effects in high-dimensional data. 
Similarly, Jin et al.\cite{jin2023cladder} introduced CAUSALCOT, which applies back-door adjustment to improve causal reasoning and has achieved promising results. 
Temp-adapter\cite{chen2023tem} proposed a contrastive learning approach to align driving videos and text at both temporal and semantic levels, significantly improving the model's scene understanding capabilities. 
To be different, we employ a feature disentanglement method to establish direct, indirect, and confounding causal relationships between scene features and behavioral features.
\subsection{Video Understanding}
Video Understanding provides users with explanations of events and their logical relationships within videos\cite{gao2021hierarchical}\cite{yan2023prompt}, delivering cohesive and abstract information\cite{li2023learning}. 
Numerous methods based on LLMs have been developed to enhance video understanding in complex environments.
For instance, Ataallah et al.\cite{ataallah2024minigpt4} proposed MiniGPT4-Video, a model that processes image and text tokens to align interactive content and projects them into the LLMs space, thereby improving video comprehension.
Xu et al.\cite{xu2024drivegpt4} developed DriveGPT4, a multimodal LLM that leverages multimodal data to interpret scenes and enhances understanding through human interaction. 
Additionally, Yuan et al.\cite{yuan2024rag} presented RAG-Driver, which integrates Retrieval-Augmented Generation (RAG) to identify similar scenes, thereby boosting the accuracy of video understanding. 
While these approaches leverage the extensive memory and computational power of large-scale LLMs\cite{dinh2024trafficvlm}\cite{malla2023drama} to achieve SOTA performance, MCAM adopts a more efficient and compact architecture with significantly fewer parameters.

\section{Methodology}
\begin{figure*}[htbp]
	\centering  
		\begin{minipage}[b]{0.75\textwidth}  
			\includegraphics[width=\textwidth]{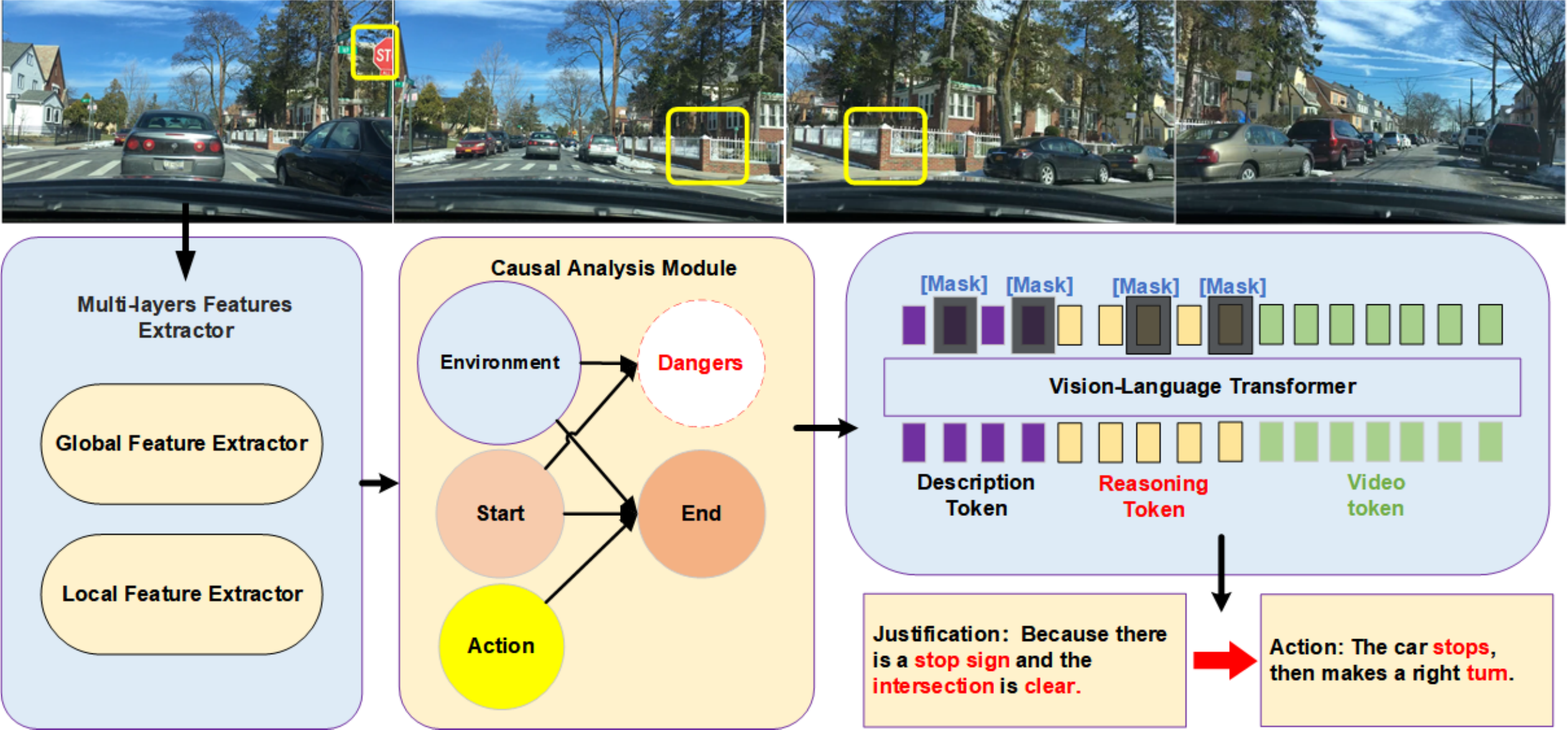}  
	\end{minipage} 

	\caption{Overview of the MCAM. The MCAM framework consists of three key components: the MFE module, which extracts multi-level features from the video clip frames; the CAM, which builds and analyzes causal relationships within the extracted key features; the VLT module, which generates coherent and contextually relevant captions by integrating visual and textual information. The yellow boxes are added by manually to support the observation results.}  
	\label{fig2}  
\end{figure*}
\vspace{-5pt}
The framework of the MCAM is shown in Fig~\ref{fig2}, which is an ego-vehicle visual driving video understanding architecture. In this section, we present the detailed implementations of MCAM.

\subsection{Multi-level Feature Extractor}
Conventional feature learning methods predominantly rely on Transformer-based architectures, which excel in capturing global dependencies but are relatively less efficient at extracting local features. To address this, we propose an efficient Multi-level Feature Extractor (MFE) that simultaneously captures both global and local features. This approach enhances the model's ability to understand complex patterns through multi-level feature representation.

In the Multi-level Feature Extractor, we utilize 3DResNet for local information extraction, and VidSwin for global information extraction. Given raw video frames of size $B \times F \times H \times W \times 3$, where $B$ is the batch size, $F$ is the number of frames, and each frame has dimensions $H \times W \times 3$, we feed the input to both the VidSwin and 3DResNet to extract global and local features, respectively. Then the features will be refined for size  $B \times \frac{F}{2} \times \frac{H}{32} \times \frac{W}{32} \times 8C$, and after the windows setting, the feature size change to $B\times \frac{FC}{2} \times \frac{H \times W}{32 \times 32}$, which $C$ equals 64. The feature from the local features extractor will be $B \times F \times \frac{H}{16} \times \frac{W}{16} \times 16C$, to keep the consistency of features, we set the down-sampling times as 4, adjust the channels by the $1 \times 1$ convolution, and fused the features by linear layer.

\subsection{Causal Analysis Module}
Existing methods are inadequate in addressing spurious causal interference in driving video understanding, as they neglect the importance of modeling the ego vehicle's state. Therefore, we propose a causal analysis module to tackle this issue. Following Pearl et al.\cite{pearl2009causality},  we consider the status transformation graph $G$, consisting of a start safe status $X_s$ and the unsafe status $W$, the action $Y$, the end safe status $X_e$, and changed environment $Z$. The safety state generally denotes the state of motion or stillness, encompassing relatively stable conditions such as stopped and uniform motion. Actions include behaviors that alter the vehicle's state, such as accelerating, decelerating, braking, lane changing, turning, and reversing. The environment encompasses weather, surrounding vehicles, traffic signals, pedestrians, obstacles, and all other elements that could impact vehicle operation. An unsafe state denotes a foreseeable collision or fall that could result if the vehicle's current driving state remains unchanged due to environmental changes.

\begin{figure*}[htbp]
	\centering  
		\begin{subfigure}[b]{0.45\textwidth}
			\hspace{2.5cm}
			\includegraphics[width=0.33\textwidth]{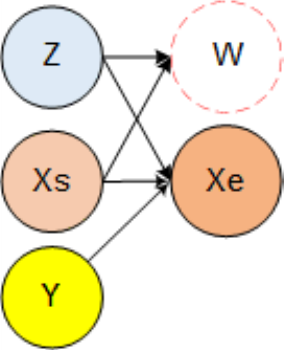}
			\caption{}
			\label{fig3.a}
		\end{subfigure} 
		\begin{subfigure}[b]{0.45\textwidth}  
			\hspace{-0.5cm}
			\includegraphics[width=1.1\textwidth]{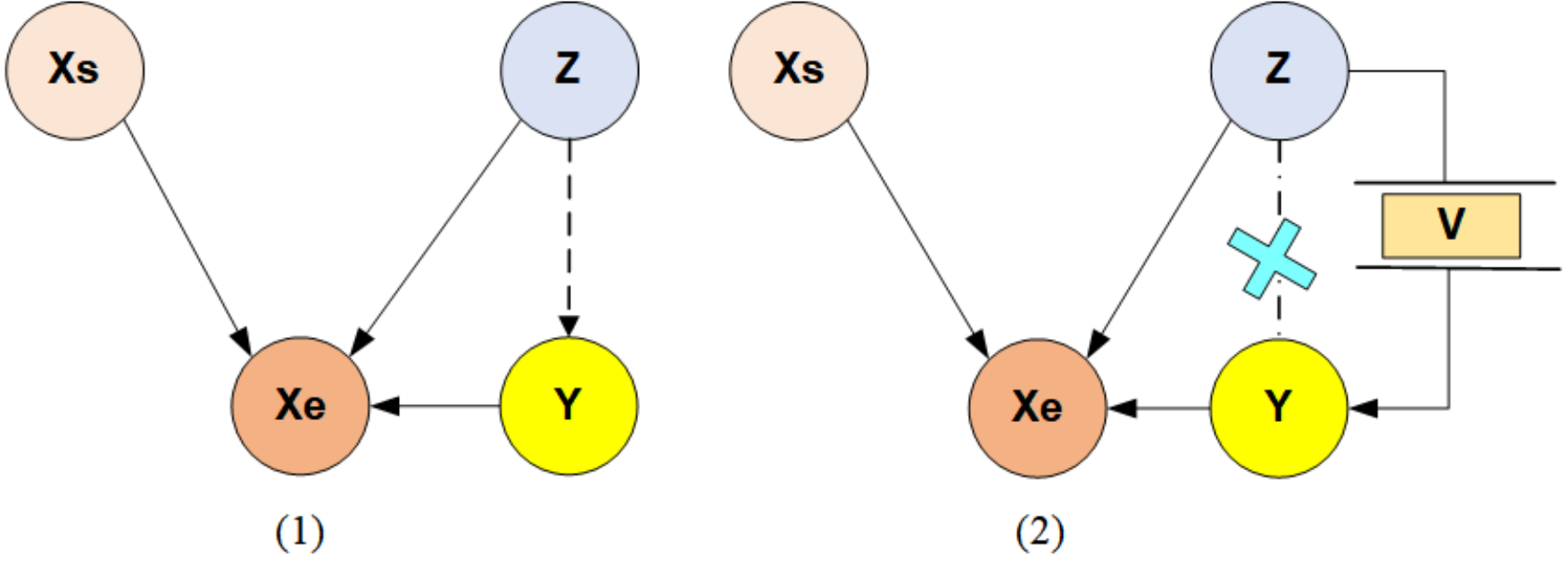} %
			\caption{}
			\label{fig3.b}
		\end{subfigure} 
	  \caption{the (a) is Driving Status Directed Acylics Graph (DSDAG), and (b) is the Refined and involved causal graph from DSDAG.}
	  \label{fig3}
\end{figure*}

Furthermore, we construct the driving status causal graph, which comes from the DSDAG, as shown in Fig \ref{fig3.a}. In this graph, $X_s$ represents the start status, $Y$ denotes the action of the ego vehicle, $X_e$ represents the end status, and $Z$ represents the changed environment respectively. The following is hypothesis and defined based on the preceding discussion:

\textbf{Driving Action} $(Y)$: This include the general operation when the ego car driving in the urban road.
\begin{equation}
	Y =\{accelerating, decelerating, turning,  \ldots \} 
\end{equation}

\textbf{Driving Environment} $(Z)$: This include the traffic condition of road which ego car stands .
\begin{equation}
	Z =\{weather, road\,type, congestion,  \ldots \} 
\end{equation}

\textbf{Vehicle State}
$(U)$: This encompasses the current state of the vehicle, including:
\begin{equation}
	U=\{speed, direction, loading\,condition, \ldots \} 
\end{equation}

\textbf{Safe Driving State} ($U_{v}$): Safe driving status is determined by the combination of the vehicle's state $U_s$ and the environmental conditions $Z_s$: 
\begin{equation} 
	U_v = \{ U_s, Z_s \}
\end{equation}

\textbf{Initial Safe Driving State}
$(X_s)$, and \textbf{Ended Safe Driving State} $(X_e)$: They are the general situation in the safe driving state.
\begin{equation} 
	\{X_s, X_e\} \in U_s
\end{equation}

\textit{Hypothesis 1}: The environment \( Z \) remains invariant at the instants corresponding to the initial state \( X_s \) and the final state \( X_e \). However, during the vehicle's transition from \( X_s \) to \( X_e \), the environment undergoes gradual and predictable changes.  

\textit{Hypothesis 2}: Throughout the transition from the initial state \( X_s \) to the final state \( X_e \), the vehicle executes a sequence of actions \( Y \). The change in the vehicle's state is jointly driven by the environment \( Z \) and the actions \( Y \).  

\textit{Hypothesis 3}: The state transition graph encodes specific causal relationships through its nodes and edges. Specifically, the action \( Y \) is determined by the current state \( X_s \) and the environment \( Z \), while the new state \( X_e \) is influenced by the action \( Y \) and the environment \( Z \).  

We posit that the various elements with same kind in the environment are linearly related. Defined the $\Gamma=\{\gamma_1, \gamma_2, \gamma_3,\dots, \gamma_n \} $ in the overall driving environment, each $\gamma_i$ represents the key coefficient of factors during vehicle driving. 
For $ \forall \xi$, $\Gamma_i$, the specific driving environment $Z_\xi$ is defined as follows:
\begin{equation}
	Z_\xi = \{U_v, \Gamma_i, \xi\}, \xi  \perp \Gamma_i
\end{equation}
where, \(U_v\) represents the vehicle state, and \(Z_s\) can be considered as the initial environment state when $\xi=\emptyset$, and $\Gamma_i$ is part of the overall environment \(\Gamma\). The specific driving environment \(Z_\xi\) combines the vehicle state, the relevant environmental factors, and a random noise term \(\xi\) that is dependent of the environmental factors. Accordingly (Fig \ref{fig3.b}.(1)), the hidden danger state should be the result of the combined action of environmental changes and unchanged vehicle states. We define the set of dangerous states $W =\{w_1, w_2,..., w_p \}$, including collisions, falls, loss of control of vehicles, violations of road traffic regulations, etc. Here, $p$ is a finite positive integer.
The hidden danger state $W_{hidden}$ 
is determined by the function $F_W$ as follows:
\begin{equation}
	W_{hidden} = F_W(U_s, Z_{\xi}), W_{hidden} \in W
\end{equation}

The role of function $F_W$ in determining whether the initial state is compatible with the environment is self-evident.
To avoid the occurrence of dangerous state $W$, drivers will choose safety action $Y$ at the appropriate time to adjust the vehicle's state. Define the set $Y_c=\{c_1, c_2,..., c_p\}$ of $Y$, which includes simple or complex program actions such as shifting gears, changing lanes, starting, stopping, etc.
Under the influence of environment $Z_{\xi}$, initial state $X_s$ will develop towards dangerous state $W_m$, while under the action of safe driving action $Y_c$, initial state $X_s$ will turn towards safe state $X_e$. The following formula:\\
\begin{align}
	U_e &= F_Y(X_s, Z_{\xi}, Y_c) \\
	X_e &= \{U_e, Z_e\}
\end{align}
at this point, the secure state $X_e$ transitions to the next initial state $X_s$, forming a logical closed loop. 

\begin{figure}[htbp]  
	\centering  \includegraphics[width=0.5\textwidth]{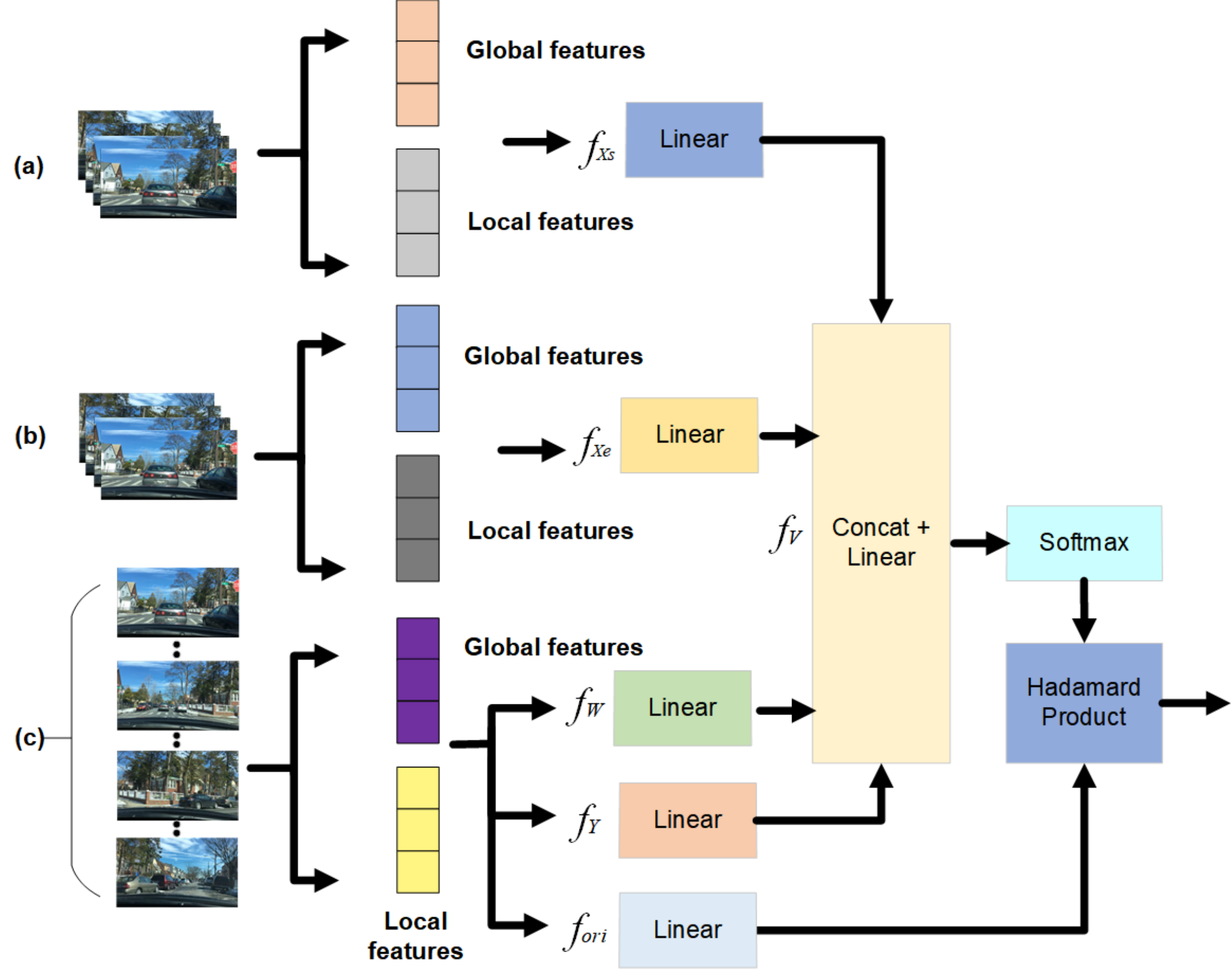}  
	\caption{Overview of Causal Analysis Module, (a) is the first and second frame of video, and (b) is the last frame of video, (c) is the whole video, all of them are fed into the MFE to get the local and global feature, then, All features will be fused in special linear layer, and absorb input to $f_V$. the $f_{ori}$ aims to protect the features as the residual layer.}  
	\label{fig4}  
\end{figure}

Based on the above, the current task is to infer the reasons for the vehicle's driving behavior. This approach differs from traditional causal reasoning, which typically infers future events based on observed phenomena. Instead, it analyzes the underlying causes based on the outcomes. This can be expressed as the following equation:
\begin{align}
	U_s &= F_{X_s}(U_v, Z_s) \\
	Y_c &= F_Y(Z_{\xi}| U_s), \xi \perp U_s  \\
	W &= F_W(Z_{\xi}| U_s, do(Y_c=\emptyset))\\
	X_e &= F_{X_e}(Z_{\xi}|U_s, do(Y_c= c))
\end{align}
where $\xi$ is random noise at the safe status $X_s$, F is computing function. As the Fig \ref{fig3.b}.(2), in the driving video understanding task, the output result is the one or several factors in the environment that have the most significant impact on the original state. So the formula is as follows:
\begin{align}
	N &= \{1,2,3,\ldots,n,\ldots,  +\infty \} \\
	Z_{\xi} &= \{\gamma_1, \gamma_2, \ldots, \gamma_n | n \in N \} \\ 
	V &= \{\gamma_x, \gamma_y, \gamma_z, \ldots\}
	\label{Eq:10}
\end{align}
when the environment $Z_s$ interacts with the initial state $X_s$ in a certain scenario and transitions to the final state $X_e$ through driving action $Y$, certain variables $V$ in the environment $Z_{\xi}$ will serve as key influencing factors for determining the appearance of $Y_c$, while the influence of other variables can be ignored. These key influencing factors $V_{\gamma}$ provide the primary basis for understanding the occurrence of action $Y_c$ occurs.
\begin{align}
	U_e &= F_{X_e}(Z_{\xi}|U_s, do(Y_c=c)) \nonumber\\
	    &= F_{X_e}(V|U_s, do(Y_c=c)) \nonumber\\
	F_V(V) &= \sum_{i=1}\sum_{j=1}{P(U_e|Y_c=c_j, V=\gamma_i)} 
\end{align}
\begin{equation}
	argmax\{F_V(V)\}
\end{equation}

where the probability of $V$ could be computed by the distribution of $V_{\gamma}$ and specific $Y_c$. During the driving status transformation process, these variables are closely related to traffic regulations, road conditions, pedestrians, and vehicles.
As shown in Fig \ref{fig4}, we implement the CAM to fusion the multi-level features as follow.

Given inputs  come from MFE:
\begin{equation}
	\begin{split}
		F_{{init}_{global}}, F_{{init}_{local}}, F_{{end}_{global}}, \\
		F_{{end}_{local}}, F_{{whole}_{global}}, F_{{whole}_{local}}
	\end{split}
\end{equation}
where $F_{init}$ is the is the initial frame of the video, used to record the initial state of the ego-vehicle, the $F_{end}$ is the ended frame of the video, used to record the ended state of the ego-vehicle.

Global-Local Fusion is performed as follows:
\begin{align}
	F_{init} &= W_{X_s}\left(F_{{init}_{global}} + F_{{init}_{local}}\right) \nonumber \\
	F_{end} &= W_{X_e}\left(F_{{end}_{global}} + F_{{end}_{local}}\right)\nonumber \\
	F_{pot} &= W_w\left(F_{{whole}_{global}}, F_{{whole}_{local}}\right) \nonumber\\
	F_{act} &= W_Y\left(F_{{whole}_{global}}, F_{{whole}_{local}}\right)\nonumber\\
	F_{ori} &= W_o\left(F_{{whole}_{global}}, F_{{whole}_{local}}\right)
\end{align}

Causal fusion combines these features:
\begin{equation}
	H = \text{Concat}(F_{init}, F_{end}, F_{pot}, F_{act})
\end{equation}

Attention mechanism computes weights $\alpha$:
\begin{equation}
	\alpha = \text{Softmax}(W_H H + b_H)
\end{equation}

The final output feature $F$ is computed by 
\begin{equation}
	F = \alpha  \odot F_{ori}
\end{equation}

\subsection{Vision-Language Transformer}
Lastly, we propose the Vision-Language Transformer (VLT) to translate causal feature relationships into language expressions. Due to the causal feature size is $B \times 4FC \times \frac{H}{32} \times \frac{W}{32}$, we tokenize the sentence, which includes both the description and the reasoning, and pad it to a fixed length. The tokenized sentence is then embedded, and a multi-layer perceptron (MLP) is used to align the causal features. To maintain the correct relationship between the video and the corresponding sentences and to prevent model hallucination, we employ the sparse attention mask\cite{lin2022swinbert}. The sparsity constraint is defined as follows:
\begin{equation}
	L_{sparse} = \lambda \times \sum^M_{i=1}(\sum^M_{j=1}(|V_{(i,j)}|)
\end{equation}
where $V$ is the relationship matrix between the causal features and word embeddings, and $M$ is the length of the causal feature vector, $\lambda$ is regularization parameter. This setting is crucial for ensuring consistency between the objects and the corresponding sentences, thus mitigating the risk of model hallucination. Additionally, the overall loss in MCAM includes signal losses (if there are any signal inputs) and sentence prediction losses. The sentence prediction losses are composed of Cross Entropy Loss for optimization, KL divergence, the $Loss_{total}$ is as follow:

\begin{align}
	L_{signal} &= \frac{1}{2N} \sum_{i=1}^{N} |y_i - \hat{y}_i| + (y_i - \hat{y}_i)^2 \nonumber \\
	L_{caption} &= -\frac{1}{N} \sum_{i=1}^{N} \sum_{c=1}^{C} y_{i,c} \log(\hat{y}_{i,c}) \nonumber \\
	&+ \beta \cdot \sum_{i=1}^{N} \sum_{c=1}^{C} P(y_{i,c}) \log \left( \frac{P(y_{i,c})}{Q(\hat{y_{i,c}})} \right)  \nonumber \\
	L_{total} &= L_{signal} + L_{caption}
\end{align}
where $N$ is the length of sequence, the $C$ is the class of words, the $y_i,\hat{y_i}$ is the sign and signal label, the $P(y_{i,c}), Q(\hat{y_{i,c}})$  is the distribution of prediction, and ground truth, respectively.

\section{Experiments}
\subsection{Datasets \& Device}
In this section, we conduct experiments on two ego-vehicle level driving understanding datasets: BDD-X \cite{kim2018textual} and CoVLA \cite{arai2024Covla}, to verify the effectiveness of MCAM and its components through comparing with existing methods.

\textbf{BDD-X Dataset.} The BDD-X dataset is an open-source driving-domain caption dataset, consisting of 7, 000 videos paired with descriptions, reasoning, and control signals. Each video is segmented into frames of $1280 \times 720$ pixels at a uniform frame rate of 30 frames per second. The dataset is meticulously annotated, with each video segment containing no more than five actions, all described in textual form. We perform uniform sampling on video clips of varying lengths (from 3 to 28 seconds), extracting 32 frames while preserving the original frame rate information to ensure temporal accuracy. We follow the existing work\cite{jin2023adapt}, and split the BDD-X dataset into 21,143, 2, 519, and 2, 859 for training, validation and testing.

\textbf{CoVLA Dataset.} The CoVLA dataset consists of 10,000 videos, each paired with detailed descriptions and control signals. The descriptions encompass critical information such as vehicle status, weather conditions, traffic signal states, pedestrian activities, and essential safety guidelines for driving. Each video spans 30 seconds, with 20 captions generated per second. In this study, six fixed timestamps were uniformly sampled from each video, along with their corresponding labels. The dataset is divided into 42, 000 video clips for training, 6, 000 clips for validation, and 12, 000 clips for testing.

\begin{table}  
	\centering 
	\caption{Experiment setting on All Datasets.} 
	\resizebox{\columnwidth}{!}{
		\begin{tabular}{c|c|c|c|c}
			\hline
			Datasets & Batch size & Training Epochs& Initial Learning Rate & Decays over time \\ \hline
			BDD-X  & 16 &   40 & 0.0003 & 16 \\ 
			CoVLA & 16 & 40 &  0.0001 & 16 \\ \hline
	\end{tabular}}
	\label{tab1}
\end{table}

\textbf{Training Details.} The MCAM is implemented using PyTorch 1.8 with CUDA 11.3. All video frames are preprocessed to a resolution of $224\times224\times3$, and the clips are sampled before training with 32 frames. Other settings are as follows Table~\ref{tab1}. We train all models on Ubuntu 24.04 using single A100 GPU with 80 GB of memory for 40 epochs. 

\textbf{Evaluation Matrics.} In this part, we evaluate MCAM through the metrics index, including BLEU-4(B4)\cite{papineni2002bleu}, METEOR(M)\cite{banerjee2005meteor}, ROUGE(R)\cite{lin2004automatic}, and CIDEr(C)\cite{vedantam2015cider}. 
\begin{figure*}[htbp]  
	\centering  
	\includegraphics[width=0.75\textwidth]{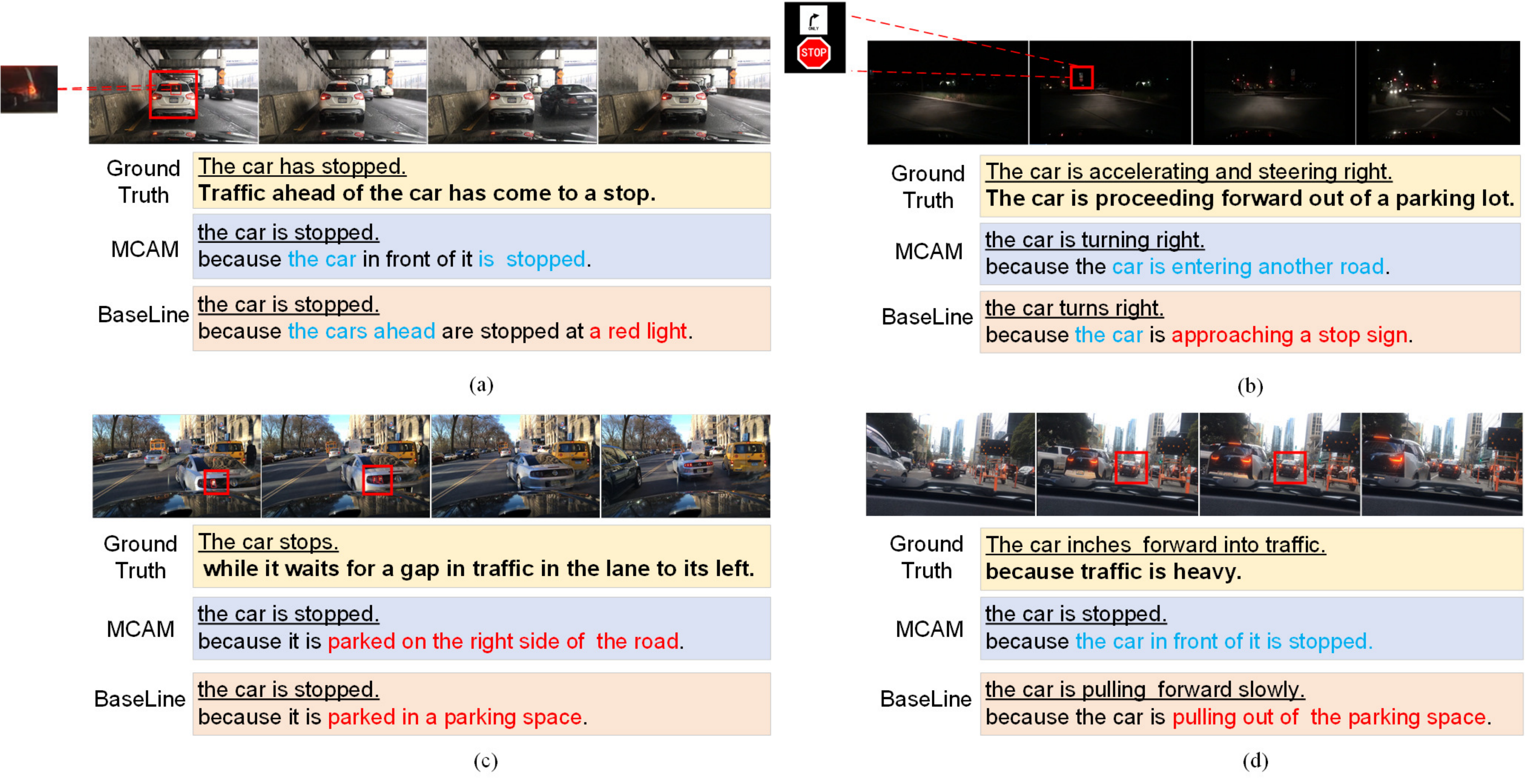}  
	\caption{Visualization for comparing the result with ADAPT in BDD-X dataset, and the red boxes are added by manually to support the observation results. The first red box highlights the taillights of the vehicle ahead of the car in front, while the second red box indicates a 'STOP' sign that requires vehicles to stop and only allows right turns.}  
	\label{fig5}  
\end{figure*}
\subsection{Main Results}
\begin{table*}  
	\centering  
	\caption{Experiment on BDD-X dataset, the narration is driving action description and reasoning is explanation.} 
	\begin{adjustbox}{max width=0.7\textwidth}
		
		\begin{tabular}{c|c|c|c|c|c|c|c|c|c|c}
			\hline
			\multirow{2}{*}{Method} & \multicolumn{4}{c|}{Narration} & \multicolumn{4}{c|}{Reasoning} & \multirow{2}{*}{Param} & \multirow{2}{*}{FPS} \\ \cline{2-9}
			& B4 & C & M & R & B4 & C & M & R & & \\ \hline
			S2VT & 30.2 & 179.8 & 27.5 & - & 6.3 & 53.4 & 11.2 & - & 86.3M & -\\ 
			SAA & 31.8 & 214.8 & 29.1 & - & 7.1 & 66.1 & 12.2 & - & 34.1M &-\\ 
			WAA & 32.3 & 215.8 & 29.2 & - & 7.3 & 69.5 & 12.2 & - & 35.7M &-\\ 
			DriveGPT4 \cite{xu2024drivegpt4} & 30.0 & 214.0 & 29.8 & - & 9.4 & 102.7 & 14.6 & -&7.85B &- \\
			RAG-Driver \cite{yuan2024rag} & 34.3 & 260.8 & 30.7 & - & 11.1 & 109.1 & 14.8 & - & 7.08B& -\\ 
			Baseline & 33.4 & 241.6 & 30.3 & 62.6 & 8.2 & 75.5 & 14.6 & 30.7 & 620.2 M &  365 \\ 
			
			MCAM(ours) & \textbf{35.7} & \textbf{252.0} & \textbf{30.5} & \textbf{63.4} & \textbf{9.1} & \textbf{94.1} & \textbf{14.0} & \textbf{32.0} & \textbf{885.3 M} & 336\\ \hline
		\end{tabular}
	\end{adjustbox}
	\label{tab2}  
\end{table*}
\textbf{Compared with SOTA methods on BDD-X.} 
As shown in Table \ref{tab2}, we compare our proposed MCAM with the state-of-art method. The ADAPT was re-implement using the same training settings and hyperparameter configurations. Additionally, we compare MCAM with the following methods: S2VT\cite{venugopalan2015sequence}, WAA, SAA \cite{kim2018textual}, RAG-Driver\cite{yuan2024rag}, DriveGPT4\cite{xu2024drivegpt4}. By leveraging the construction of the DSDAG, MCAM achieves superior accuracy on the BDD-X dataset compared to the baseline, while effectively mitigating spurious correlations. Specifically, for the narration task, MCAM achieves the B4 of 35.7\%, because the causal analysis module could construct the dependence key factors with action. In the reasoning task, our MCAM achieves a B4 metric of 9.1\%. In Fig \ref{fig5}.a, the MCAM accurately identifies the red brake lights on the vehicle and attributes the stop to the car in front. This indicates that the stop is more likely caused by the red traffic light rather than the preceding vehicle. The brake lights may interfere with the model's judgment. In Fig \ref{fig5}.b, the stop sign is hard to recognize. The baseline erroneously attributes the turning action to the sign, creating a spurious relationship. Our method, however, focuses on the driving process, particularly the transition from the parking lot to another road. In Fig \ref{fig5}.c, both the baseline and MCAM fail to recognize the ego vehicle's lane change requirement, instead misinterpreting the stopping behavior as parking intention. This occurs due to two key factors: (1) a yellow van in the front-right region is stationary while interacting with pedestrians, and (2) the vacated space from a departing leading vehicle presents a viable parking opportunity. While this interpretation diverges from ground truth, it represents a reasonable contextual inference. Notably, MCAM demonstrates superior performance in accurately localizing the parking target. 

\begin{figure*}[htbp]  
	\centering  \includegraphics[width=0.8\textwidth]{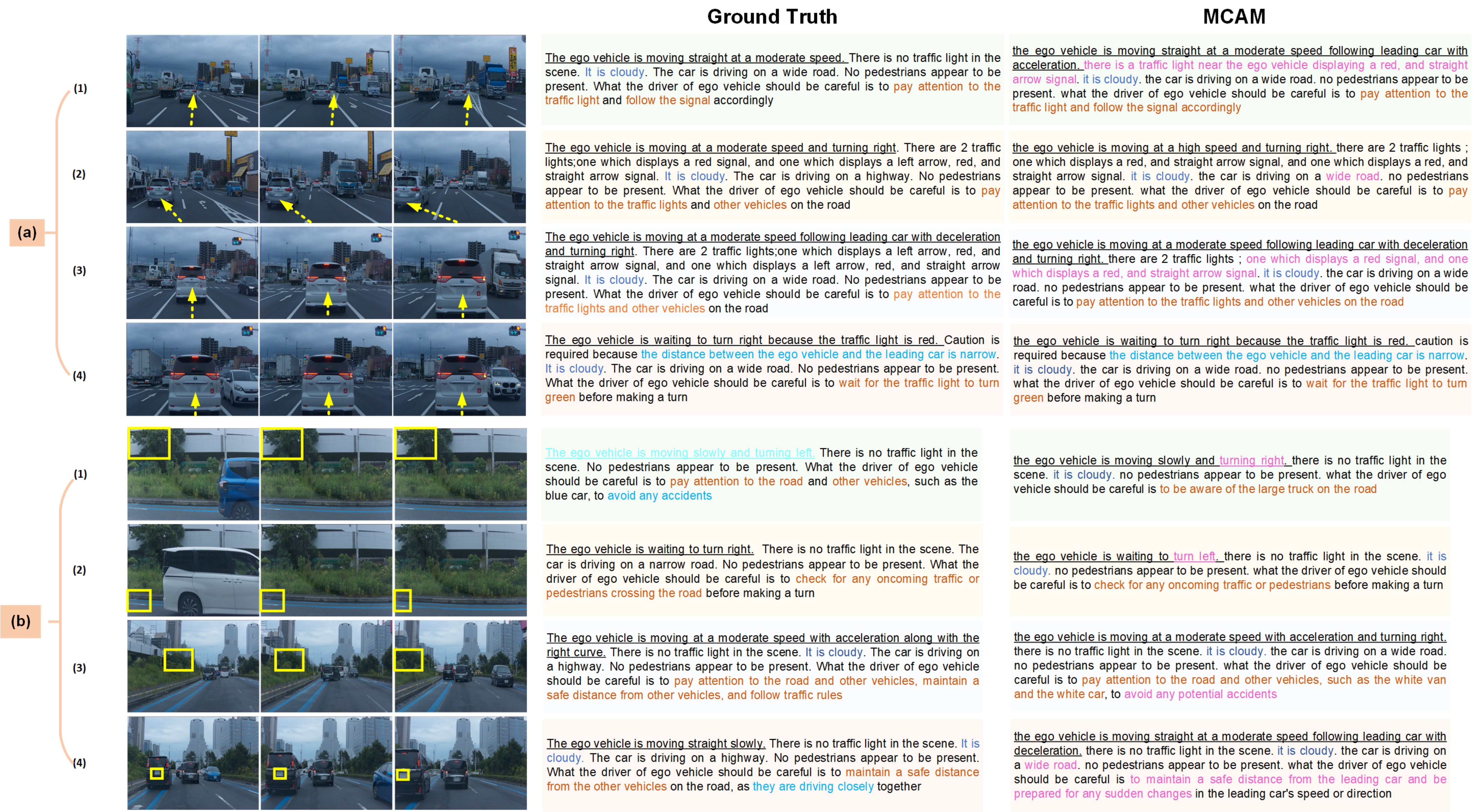}  
	\caption{The visualization of MCAM on the CoVLA dataset, includes the ground truth (GT) caption, and the result from MCAM prediction. The yellow arrows and boxes is added by manually to support the observation results, and the words with underline indicate the description of ego-car, the blue words are the weather, brown words emphasis the caution of driving, the sky blue words are the reason of caution, and the words highlighted in pink indicate the discrepancy between the ground truth and our method.}  
	\label{fig6}  
\end{figure*}

\textbf{Generation Experience on CoVLA.}
On a driving video understanding dataset of the same category, we reproduced the baseline model ADAPT and evaluated its performance using eight widely recognized metrics: BLEU-1 (B1), BLEU-2 (B2), BLEU-3 (B3), BLEU-4 (B4), METEOR (M), ROUGE (R), SIPCE(S)\cite{anderson2016spice} and CIDEr (C). As depicted in Fig \ref{fig6}, the MCAM demonstrates effective identification of ego vehicle behaviors within videos.
Specifically, as shown in Fig \ref{fig6}.a.(1), from the 1st to the 5th frames, the ego-vehicle maintains a straight trajectory, consistent with the label information. The MCAM not only accurately describes the vehicle's movement state but also exhibits robust recognition capabilities for weather conditions and traffic signals. More importantly, the model almost comprehensively covers all critical attention factors in complex traffic scenarios. In Fig \ref{fig6}.a.(2), the model accurately identifies the number of cameras and interprets their content, offering pertinent advisories for safe vehicular operation. This capability effectively demonstrates the model's inferential prowess. Regarding the identification of lane-changing behaviors, our method also performs excellently. Particularly in Fig \ref{fig6}b.(1), where labels generated by the Llama-7b model based on image understanding of specific frames incorrectly predicted the vehicle’s direction, our model, after fine-tuning, significantly corrects this misjudgment, demonstrating strong generalization ability. 
\begin{table}  
	\centering 
	\caption{The generated experiment on CoVLA dataset.} 
	\begin{adjustbox}{max width=0.5\textwidth}
		\begin{tabular}{c|c|c|c|c|c|c|c|c|c|c}
			\hline
			\multirow{2}{*}{Method} & \multicolumn{10}{c}{CoVLA} \\ \cline{2-11}
			& B1 & B2& B3 & B4 & C & M & R & S & Para & FPS\\  \hline
			Baseline & 81.9&78.5 &76.1&74.2& 236.9 & 48.8 & 80.7 &78.5 & 620.2M & 94.2 \\ 
			MCAM(ours) & \textbf{82.6}&\textbf{79.4}&\textbf{77.2}&\textbf{75.3}& \textbf{275.4} & \textbf{50.2} & \textbf{81.9} & \textbf{79.9}  & \textbf{885.3M}& \textbf{86.22}\\ \hline
		\end{tabular}   
		\label{tab4}  
	\end{adjustbox}
\end{table}

\begin{table}  
	\centering  
	\caption{Ablation experiment on BDD-X dataset.} 
	\begin{adjustbox}{max width=0.5\textwidth}
		\begin{tabular}{l|c|c|c|c|c|c|c|c|c|c}
			\hline
			\multirow{2}{*}{Method} & \multicolumn{4}{c}{Narration} & \multicolumn{4}{c|}{Reasoning} & \multirow{2}{*}{Param} & \multirow{2}{*}{FPS}\\ \cline{2-9}
			& B4 & C & M & R & B4 & C & M & R & & \\\hline
			VidSwin + VLT & 32.9 & 235.8 &  29.8 & 62.2 &8.0 & 74.3 & 14.6 & 31.1 & 582.2 M & 374\\ 
			
			3DResNet + VLT&  32.8  & 221.8 & 28.97&  61.4 & 6.4 & 55.3 & 12.5 & 28.1 & 494.0M & 616\\
			
			MFE + VLT & 33.4  &  218.1 & 29.5 &  61.6& 7.1 & 79.8 & 12.8 & 30.5 & 845.8 M &  364\\ 
			
			VidSwin +CAM+ VLT & 34.2 & 242.7 &  30.0 & 62.5 & 8.2 & 80.3 & 14.0 & 31.1 & 588.3 M & 336\\ 
			MFE + CAM + VLT (MCAM) & \textbf{35.3} & \textbf{251.6} & \textbf{30.7} & \textbf{63.1} & \textbf{9.0} & \textbf{92.9} & \textbf{13.8} & \textbf{31.7} & \textbf{885.3} M & \textbf{320}\\ \hline
		\end{tabular}  
	\end{adjustbox}
	\label{tab5}  
\end{table}
\subsection{Ablation Study}
As is shown in Table \ref{tab5}, we implement the ablation experiment from the caption including narration and reasoning, to prove the reasonable combination of MFE, and CAM along VLT. For the video understanding task, an accurate description of driving actions forms the foundation. By analyzing the observed actions and key influential objects, the cause of the actions can be determined more precisely and convincingly. The 3DResNet paired with VLT could get lower performance than VidSwin with VLT due to lack of global information. When the MFE encounters the VLT, narration, and reasoning performance has increased. The testing process of MFE is more steady than the single encoder, the accuracy in testing will be consistency with more steps training. Differently, including CAM, the combination of VidSwin and VLT could create a more accurate relation between the observed object and cause. Interestingly, the CAM could simultaneously indicate the narration and reasoning efficiently.

\section{Conclusion}
In this paper, we introduce the Multimodal Causal Analysis Model (MCAM) for modeling causality in ego-vehicle scenarios. MCAM incorporates a causal analysis module to enable effective identification of diverse driving behaviors under dynamic road conditions. This integration enhances the precision of predictions and interpretations of ego-vehicle behaviors. We validate the effectiveness of our approach on two ego-vehicle level benchmarks datasets BDD-X and CoVLA, demonstrating MCAM's superior performance and broad applicability in real-world scenarios.

\textbf{Limitation and Future Work.} Our work also has certain limitations. For example, the datasets existing incorrect labels will impact model's the recognized capability. Future work will construct higher-quality datasets and explore internal causal relationships among various objects in videos to further improve prediction accuracy and efficiency.
\vspace{-10pt}
\section*{Acknowledgements}
This work is supported by the National Natural Science Foundation of China (Grant No. 62472055 and No. 62172064).
{
    \small
    \bibliographystyle{ieeenat_fullname}
    \bibliography{main}
}

\section{Appendix}

\begin{figure*}[htbp]
	\centering
	\includegraphics[width=1.1\textwidth]{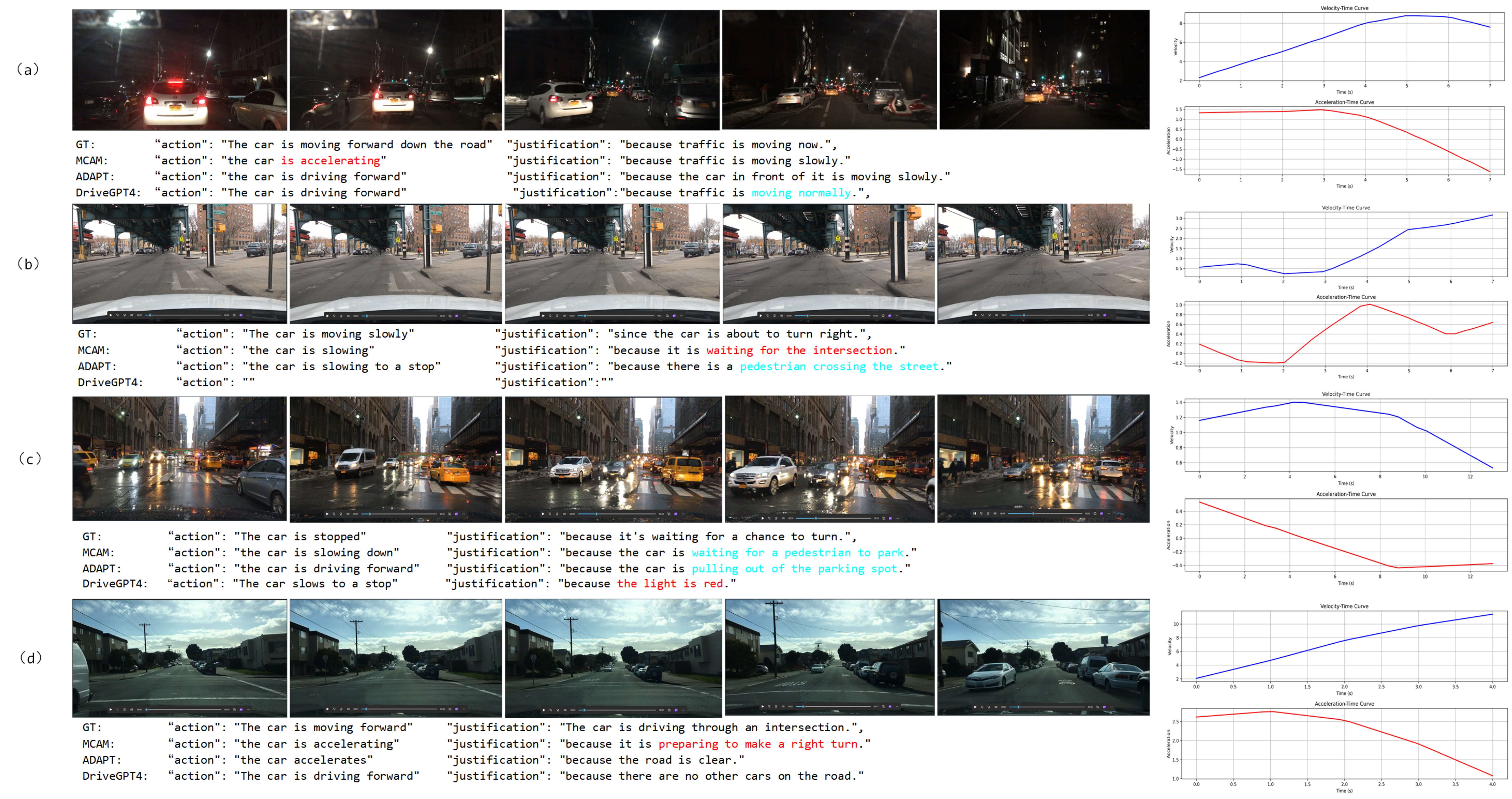}
	\caption{Four representative counterexamples demonstrate the performance disparity between DriveGPT4, ADAPT, and MCAM against Ground Truth (GT) trajectories. Annotations: red text = different expression; blue text = erroneous predictions; red curve = acceleration profile (m/s²); blue curve = velocity trajectory (m/s).}
	\label{fig10}
\end{figure*}
\textbf{Comparing in BDD-X for failure examples}: In Figure~\ref{fig10}.(a), the vehicle transitions from a stopped state to forward motion, as indicated by the speed and acceleration curves. MCAM specifically investigates the acceleration dynamics during vehicle launch phases and their propagation to straight-line driving behavior—a critical initial condition not adequately captured in conventional GT annotations. In Figure~\ref{fig10}.(b), MCAM correctly identifies the deceleration as preparatory behavior for intersection approach, demonstrating semantic alignment with ground truth labels despite lexical variations. ADAPT erroneously associates this maneuver with pedestrian avoidance but pedestrians had already cleared the crossing zone. DriveGPT4 exhibits null output instances, likely resulting from selective response filtering during evaluation. For Figure~\ref{fig10}.(c), MCAM associates the vehicle’s slowing down with a preceding pedestrian. Although visibility on the left is limited, this interpretation is more reasonable. Finally, in Figure~\ref{fig10}.(d), MCAM infers preparation for a right turn. This may result from the large gap ahead being perceived as a right-turn lane.\\
\textbf{Experiments in DRAMA}
As shown in Table \ref{tab1}, MCAM achieves satisfactory performance, outperforming LCP by 1.7\% in B1, 7.7\% in B4, 6.2\% in Meteor, 5.5\% in Rouge,  and 5.7\% in Spice.
\begin{table}   
	\centering 
	\caption{The generated experiment on DRAMA dataset.} 
	\begin{adjustbox}{max width=0.5\textwidth}
		\begin{tabular}{l|c|c|c|c|c}
			\hline
			\multirow{2}{*}{Method} & \multicolumn{5}{c}{DRAMA} \\ \cline{2-6}
			& B1 & B4 & M & R & S \\  \hline
			LCP\cite{malla2023drama} & 73.9 & 54.7 &39.1& 70.0 & 56.0 \\ 
			ADAPT & 74.5 &60.8 & \textbf{45.5} & 75.3 & 60.1\\
			MCAM(ours) &\textbf{75.8} & \textbf{62.4} &45.3 & \textbf{75.5} & \textbf{61.7} \\ \hline
		\end{tabular}   
		\label{tab1}
	\end{adjustbox}
	\vspace{-15pt}
\end{table}

\end{document}